\documentclass[10pt,twocolumn,letterpaper]{article}

\usepackage{cvpr}
\usepackage{times}
\usepackage{epsfig}
\usepackage{graphicx}
\usepackage{amsmath}
\usepackage{amssymb}
\usepackage{algorithm}
\usepackage{algorithmic}
\usepackage{multirow}
\usepackage{caption}
\usepackage[pagebackref=true,breaklinks=true,colorlinks,bookmarks=false]{hyperref}

\cvprfinalcopy 
\def\cvprPaperID{717} 

\ifcvprfinal\fi
\begin{document}

\title{Dynamic Video Segmentation Network}
\author{Yu-Syuan Xu, Tsu-Jui Fu\thanks{Equal contribution}, Hsuan-Kung Yang\footnotemark[1]\textit{, Student Member, IEEE} and Chun-Yi Lee\textit{, Member, IEEE}\\
Elsa Lab, Department of Computer Science, National Tsing Hua Uiversity\\
{\tt\small \{yusean0118, rayfu1996ozig, hellochick\}@gapp.nthu.edu.tw, cylee@cs.nthu.edu.tw}
\vspace{-15pt}
}
\maketitle
\thispagestyle{empty}

\begin{abstract}
In this paper, we present a detailed design of dynamic video segmentation network (DVSNet) for fast and efficient video semantic segmentation.  DVSNet consists of two convolutional neural networks: a segmentation network and a flow network.  The former generates highly accurate semantic segmentations, but is deeper and slower.  The latter is much faster than the former, but its output requires further processing to generate less accurate semantic segmentations.  We explore the use of a decision network to adaptively assign different frame regions to different networks based on a metric called expected confidence score.  Frame regions with a higher expected confidence score traverse the flow network.  Frame regions with a lower expected confidence score have to pass through the segmentation network.  We have extensively performed experiments on various configurations of DVSNet, and investigated a number of variants for the proposed decision network.  The experimental results show that our DVSNet is able to achieve up to 70.4\% mIoU at 19.8 fps on the Cityscape dataset.  A high speed version of DVSNet is able to deliver an fps of 30.4 with 63.2\% mIoU on the same dataset.  DVSNet is also able to reduce up to 95\% of the computational workloads.
\end{abstract}

\section{Introduction}
\label{Introduction}
Fast and accurate semantic segmentation has been a fundamental challenge in computer vision.  The goal is to classify each pixel in an image into one of a given set of categories.  In recent years, image semantic segmentation has achieved an unprecedented high accuracy on various datasets~\cite{PASCAL,Cityscapes,ADE20K,COCO} via the use of deep convolutional neural networks (DCNNs)~\cite{DeepLabv1,DeepLabv2,DeepLabv3,PSPNet,ResNet-38,RefineNet,DUC,LRR}.  Accurate semantic segmentation enables a number of applications which demand pixel-level precision for their visual perception modules, such as autonomous vehicles~\cite{CARLA}, surveillance cameras, unmanned aerial vehicles (UAVs), and so on.  However, due to their real-time requirements, these applications typically require high frame rates per second (fps), necessitating short inference latency in the perception modules.  Unfortunately, contemporary state-of-the-art CNN models usually employ deep network architectures to extract high-level features from raw data~\cite{Alex,VGG,Res,Google,Googlev4}, leading to exceptionally long inference time.   The well-known models proposed for image semantic segmentation, including fully convolutional networks (FCN)~\cite{FCN}, DeepLab~\cite{DeepLabv1,DeepLabv2,DeepLabv3}, PSPNet~\cite{PSPNet}, ResNet-38~\cite{ResNet-38}, RefineNet~\cite{RefineNet}, dense upsampling convolution (DUC)~\cite{DUC}, etc., are not suitable for real-time video semantic segmentation due to their usage of deep network architectures. These models usually incorporate extra layers for boosting their accuracies, such as spatial pyramid pooling (SPP)~\cite{DeepLabv2,DeepLabv3,PSPNet,SPP}, multi-scale dilated convolution~\cite{DeepLabv1,DeepLabv2,DeepLabv3,DUC,Dilation}, multi-scale input paths~\cite{DeepLabv2,Piecewise,ICNet,Scale}, multi-scale feature paths~\cite{DeepLabv1,RefineNet,LRR,FRRN,MultiPath}, global pooling~\cite{DeepLabv3}, and conditional random field (CRF)~\cite{DeepLabv1,DeepLabv2,Piecewise,CRF2,CRF1}.  These additional layers consume tremendous amount of computational resources to process every pixel in an image, leading to impractical execution time.  In the past decade, video semantic segmentation focusing on reducing the inference time has received little attention~\cite{Clock, DFF}.  With increasing demand for high accuracy and short inference time, a method for efficiently reusing the extracted features in DCNNs for video semantic segmentation is becoming urgently necessary.

\begin{figure}[t]
\centering
\captionsetup{font=small}
\includegraphics[width=0.45\textwidth]{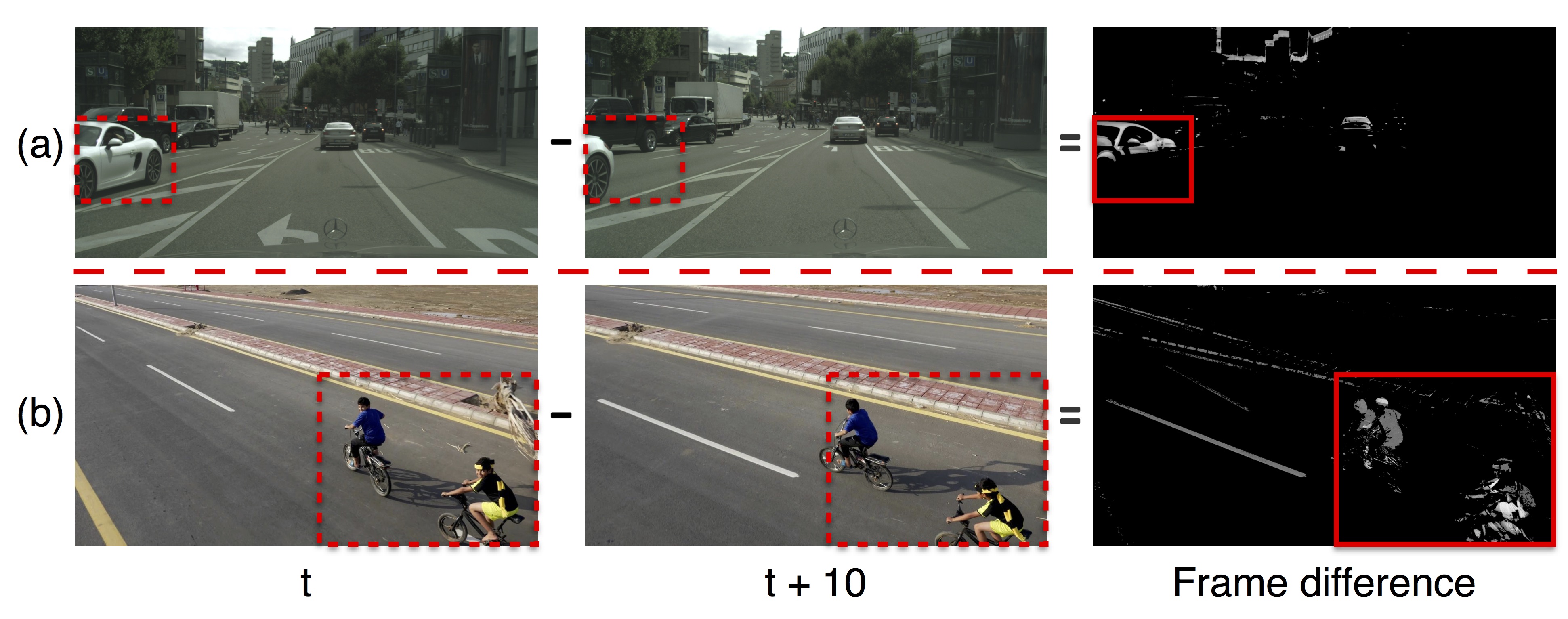}
\vspace{-8pt}
\caption{Comparison of frames at timestamps $t$ and $t+10$ in two video sequences} 
\vspace{-17pt}
\label{regional}
\end{figure}

It is unnecessary to reprocess every single pixel of a frame by those deep semantic segmentation models in a video sequence.  When comparing the difference between two consecutive frames, it is common that a large portion of them is similar.  Fig.~\ref{regional} illustrates an example of the above observation.  The left parts show the video frames at timestamps $t$ and $t+10$, respectively.  The right part shows the difference between these two frames.  It can be observed that only a small portion of the frames are apparently different (highlighted by red rectangles), implying that the a large portion of the feature maps between these frames is invariant, or just varies slightly.  Therefore, performing complex semantic segmentation on the entire video frame can potentially be a waste of time.  By keeping or slightly modifying the feature maps of the portions with minor frame differences while performing semantic segmentation for the rest, we may achieve a better efficiency and shorter latency in video semantic segmentation than per-frame approaches.

\begin{figure}[t]
\centering
\captionsetup{font=small}
\includegraphics[width=0.45\textwidth]{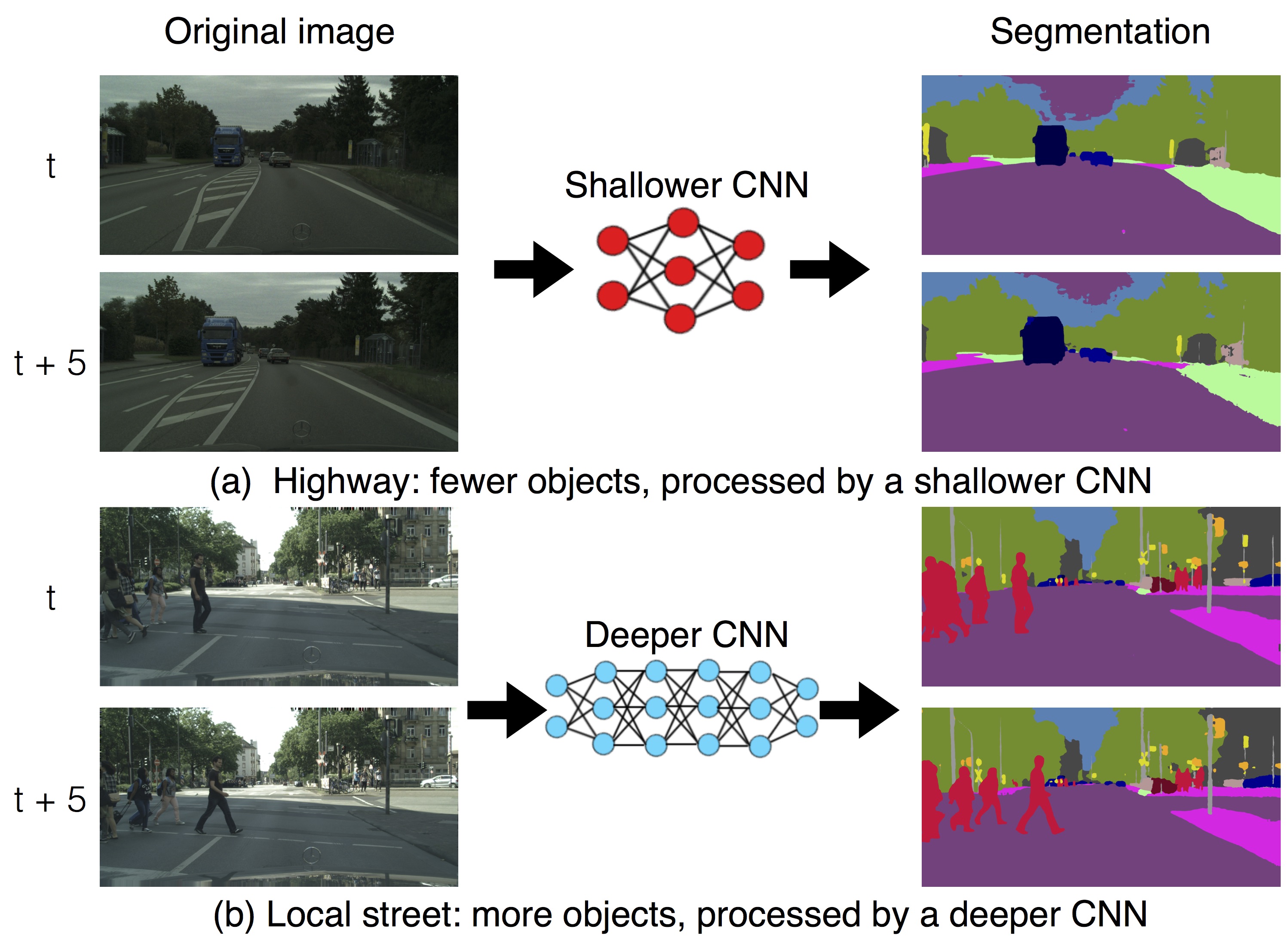}
\vspace{-7pt}
\caption{Using different CNNs for different video scenes} 
\vspace{-18pt}
\label{dynamic}
\end{figure}

Another perspective for accelerating the processing speed of video semantic segmentation is by leveraging the temporal correlations between consecutive frames.  Consecutive video frames that do not change rapidly have similar high-level semantic features~\cite{Slow1,Slow2}.  On the other hand, frames containing multiple moving objects demonstrate disparate feature maps at different timestamps.  Fig.~\ref{dynamic} illustrates an example of such scenarios.  Fig.~\ref{dynamic}~(a) shows a semantic segmentation performed on a highway, which contains fewer objects and thus results in less changes in consecutive segmented images.  Fig.~\ref{dynamic}~(b), on the contrary, corresponds to a video sequence taken from a local street, which contains dozens of moving objects.  The former suggests reusing the extracted features and updating them with as few computations as possible (e.g., by a shallower CNN), while the latter requires performing highly accurate semantic segmentation on every single frame (e.g., by a deeper CNN).  Researchers have attempted several methodologies to reuse the extracted features, such that not all of the frames in a video sequence have to traverse the entire DCNN.  The authors in~\cite{Clock} proposed to update high-level features less frequently, taking advantage of temporal correlations between frames to reduce the computation time.  Another effective approach is called Deep Feature Flow (DFF)~\cite{DFF}, which suggests to propagate the features of a few key frames to  other timestamps via the use of optical flow~\cite{Flow}.  DFF combines the concepts of segmentation network and FlowNet~\cite{FlowNet}, and is able to reduce the computation time of video semantic segmentation by 74\%.  Unfortunately, DFF uses a fixed key frame scheduling policy (i.e., it assumes a fixed update period between two consecutive key frames), which sacrificies its flexibility and customizability.

Based on the above observations, we propose a new network architecture, called dynamic video segmentation network (DVSNet), to adaptively apply two different neural networks to different regions of the frames, exploiting spatial and temporal redundancies in feature maps as much as possible to accelerate the processing speed.  One of the networks is called the segmentation network, which generates highly accurate semantic segmentations, but is deeper and slower.  The other is called the flow network. The flow network is much shallower and faster than the the segmentation network, but its output requires further processing to generate estimated semantic segmentations (which might be less accurate than the ones generated by the segmentation network).  The former can be implemented by any of the contemporary state-of-the-art architectures~\cite{DeepLabv2,PSPNet}, while the latter is developed on top of FlowNet 2.0~\cite{FlowNet2}. We divide each frame into multiple regions.  Regions with minor differences between consecutive frames, where most of the image contents are similar, should traverse the flow network (Fig.~\ref{dynamic}~(a)).  Regions with huge differences between consecutive frames, where the contents change significantly, have to pass through the segmentation network (Fig.~\ref{dynamic}~(b)).  In other words, different regions in a frame may traverse different networks of different lengths when they are presented to DVSNet.  We designate the the regions processed by the segmentation network and flow network as the key frame regions and spatial warping regions, respectively.  In order to accelerate the processing speed, we assume that key frame regions are relatively sparser than the spatial warping regions in a video. DVSNet offers two major advantages. First, efficiency is enhanced because DVSNet adapts its throughput to the differences between consecutive frame regions at runtime.  Second, significant computation can be saved by the use of the flow network.  This scheme is primarily targeted at video semantic segmentation.

To define a systematic policy for efficiently assign frame regions to the two networks while maintaining flexibility and customizability, we further propose two techniques: (i) adaptive key frame scheduling policy, and (ii) decision network (DN).  Adaptive key frame scheduling policy is a technique for determining whether to process an input frame region by the segmentation network or not.  Differing from the fixed key frame scheduling policy employed in~\cite{DFF}, the proposed adaptive key frame scheduling policy updates the key frames according to a new metric called \textit{expected confidence score}. An \textit{expected confidence score} is evaluated for each frame region.  The higher the \textit{expected confidence score} is, the more likely the segmentation generated by the flow network will be similar to that of the segmentation network.  The value of the \textit{expected confidence score} reflects the confidence of the flow network to generate similar results as the segmentation network.  The larger the frame region difference is, the more likely the flow network is unable to infer the correct frame region segmentation.  Fig.~\ref{architecture} illustrates such a scenario, which is explained in Section~\ref{Dynamic Video Segmentation Network}.  A frame region is first analyzed for its \textit{expected confidence score}. If its \textit{expected confidence score} is higher than a predefined threshold, it is processed by the flow network.  Otherwise, it is allocated to the segmentation network.  The decision threshold is adjustable for different scenarios.  A higher threshold leads to a higher mean intersection-over-union (mIoU) accuracy of the segmentation results, while a lower thresholds corresponds to a shorter processing latency for most of the frame regions.  An adjustable threshold allows DVSNet to be applied to various scenarios.  The function of DN is to determine whether an input frame region has to traverse the segmentation network by estimating its \textit{expected confidence score}.  In DVSNet, DN is implemented as a CNN, with its network size much smaller than typical network architectures for image recognition.  DN can be trained by supervised learning, where the details is covered in Section~\ref{DN and Its Training Methodology}.  With the use of DN, we are able to obtain fine-grained control of fps, and enhance the efficiency of computation for video semantic segmentation.  

To verify the proposed DVSNet, we have extensively performed experiments on several well-know models for the segmentation network, and investigated a number of variants of the proposed DN scheme.  
The results show that our method is able to achieve up to 70.4\% mIoU at 19.8 fps on the Cityscape~\cite{Cityscapes} dataset.  A high speed version of DVSNet achieves 30.4 fps with 63.2\% mIoU on the same dataset.  Our model is able to reduce up to 95\% of the computational workloads.  The contributions of this work are as follows:
\begin{enumerate}
\vspace{-4pt}
\itemsep=-3pt
\item A frame division technique to apply different segmentation strategies to different frame regions for maximizing the usage of video redundancy and continuity.
\item A DN for determining whether to assign an input frame region to the segmentation network, and adaptively adjusting the update period of the key frames.
\item An adaptive key frame scheduling policy based on a metric called \textit{expected confidence score}.
\item An adjustable threshold for DN.
\item A comprehensive analysis of the impact of DN's decision threshold on DVSNet's accuracy and fps.
\vspace{-4pt}
\end{enumerate}

  The remainder of this paper is organized as follows.  Section~\ref{Background} introduces background material.  Section~\ref{DVSNet} walks through the proposed DVSNet architecture, its implementation details, and the training methodologies.  Section~\ref{Experiments} presents the experimental results.  Section~\ref{Conclusion} concludes.
  
\section{Background}
\label{Background}
In this section, we introduce background material.  We first provide an overview of semantic segmentation and optical flow.  Then, we briefly review related work that focuses on video semantic  segmentation.

\subsection{Image Semantic Segmentation}
\label{Image Semantic Segmentation}
Various techniques have been proposed in the past few years to transfer DCNNs from image classification~\cite{Alex,VGG,Res,Google,BN} to image semantic segmentation tasks~\cite{DeepLabv3,PSPNet,ResNet-38,RefineNet,FCN}.  Fully convolutional network (FCN)~\cite{FCN} was the pioneer to replace fully-connected layers with convolutional layers.  Inspired by FCN, a number of successive methods and modifications were proposed to further improve the accuracy.  The authors in~\cite{DeepLabv1,DeepLabv2,DeepLabv3,Dilation} investigated the use of dilated (atrous) layers to replace deconvolutional layers~\cite{Deconvolution} in dense prediction tasks, so that the resolution of feature maps can be explicitly controlled.  Integrating DCNNs with a fully-connected conditional random field (CRF) have also been explored in~\cite{DeepLabv1,DeepLabv2,CRF2} to refine the image semantic segmentation results.  Spatial pyramid pooling~\cite{SPP} and atrous spatial pyramid pooling (ASPP)~\cite{DeepLabv2,DeepLabv3} are employed by PSPNet~\cite{PSPNet} and DeepLab~\cite{DeepLabv2,DeepLabv3} to capture multi-scale context information.  These methods are effective in increasing the accuracy, however, sometimes at the expense of longer latency and more computational workloads.
  
\subsection{Optical Flow}
\label{Optical Flow}
Optical flow estimation has been the subject of a number of research works~\cite{FlowNet, Flow, SDFlow, LDFlow, DeepFlow, FlowFields, PCAFlow}.  
Unfortunately, most of the previous methodologies are mainly developed for running on CPUs, failing to incorporate execution efficiency offered by GPUs.  For deep learning approaches running on GPUs, FlowNet~\cite{FlowNet} is the first model to apply DCNNs to optical flow estimation.  It then later evolves into two recent architectures.  One is called spatial pyramid network (SpyNet)~\cite{FlowNetP}, which uses the coarse-to-fine spatial pyramid structure of~\cite{FlowNetP2} to learn residual flow at each pyramid level.  The other is FlowNet 2.0~\cite{FlowNet2}, which introduces a new learning schedule, a stacked architecture, and a sub-network specialized on small motions to enhance flow estimation.  In this paper, we incorporate the building blocks from FlowNet 2.0 into our DVSNet for accelerating video semantic segmentation.

\begin{figure*}[t]
\centering
\captionsetup{font=small}
\includegraphics[width=\textwidth]{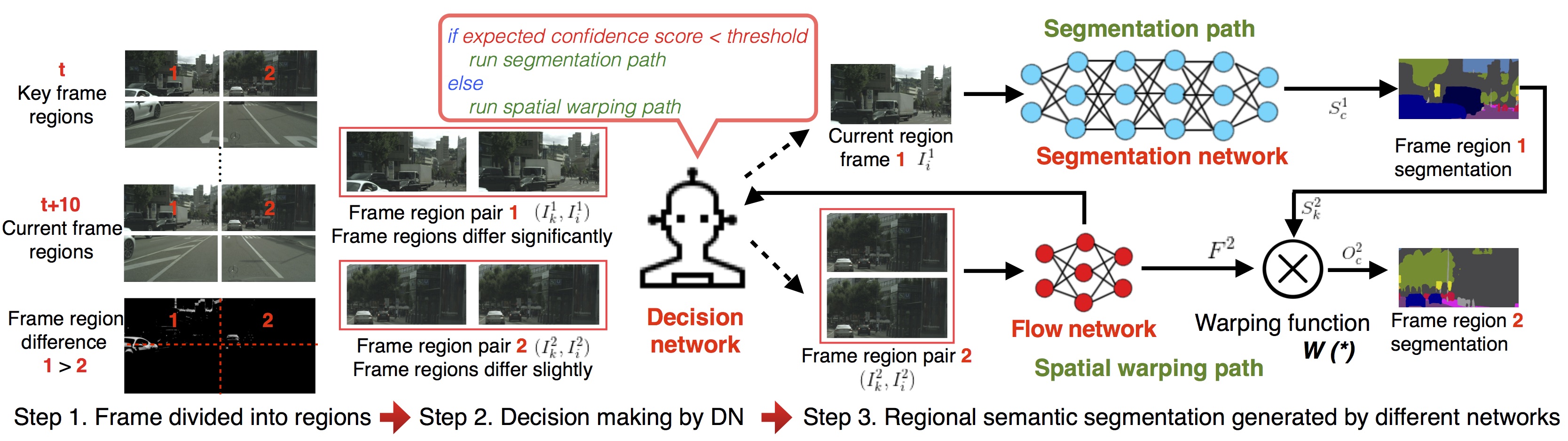}
\vspace{-20pt}
\caption{DVSNet framework} 
\label{architecture}
\vspace{-17pt}
\end{figure*}

\subsection{Video Semantic Segmentation}
\label{Video Semantic Segmentation}
Video semantic segmentation has gained researchers' attention in recent years.  Many research works are focused on efficient network architecture for speeding up video semantic segmentation~\cite{Clock, DFF}.  Clockwork employs different update periods for different layers of the feature maps in the network, and reuses feature maps of past frames in certain network layers to reduce computation~\cite{Clock}.  Deep feature flow (DFF) exploits an optical flow network to generate flow fields and propagates feature maps from key frames to nearby frames~\cite{DFF}.  It is reported that Clockwork runs 1.3$\times$ faster than per-frame approach~\cite{Clock}, however, its mIoU drops from 65.9\% to 64.4\% on the Cityscapes dataset~\cite{Cityscapes}.  In contrast, DFF runs three times faster than per-frame approach, and its mIoU only drops slightly from 71.1\% to 70.0\% on the same datasets~\cite{DFF}.  DFF demonstrates much better performance than Clockwork in both accuracy and efficiency.  However, a major drawback is that it employs a fixed key frame scheduling policy.  Inspired by DFF, the proposed DVSNet embraces an adaptive key frame scheduling policy, offering better performance than DFF in terms of both accuracy and efficiency.  

\section{DVSNet}
\label{DVSNet}
In this section, we present the architecture and implementation details of DVSNet.  We first outline the framework of DVSNet.  Next, we introduce its methodologies including adaptive key frame scheduling, frame region based execution. Finally, we explain the model of DN and its training methodology in detail.

\subsection{Dynamic Video Segmentation Network}
\label{Dynamic Video Segmentation Network}
The framework of DVSNet is illustrated in Fig.~\ref{architecture}.  The DVSNet framework consists of three major steps.  The first step in the DVSNet framework is dividing the input frames into frame regions.  In Fig.~\ref{architecture}, we assume that $I_{k}$ represents the key frame, $I_{i}$ represents the current frame, and the number of the frame regions equals four.  We further assume that the frame regions at timestamp $t$ correspond to the key frame regions, and those at timestamp $t+10$ correspond to the current frame regions. The differences between the key frame regions and the corresponding current frame regions are shown at the bottom left. In this example, region \textbf{1} shows significantly more differences in pixels between timestamps $t$ and $t+10$, while the other regions only change slightly. In step 2, DN analyzes the frame region pairs between $I_{k}$ and $I_{i}$, and evaluates the \textit{expected confidence scores} for the four regions separately.  DN compares the \textit{expected confidence score} of each region against a predetermined threshold.  If the \textit{expected confidence score} of a region is lower than the threshold, the corresponding region is sent to a segmentation path (i.e., the segmentation network).  Otherwise, it is forwarded to a spatial warping path, which includes the flow network.  The function of DN is to evaluate if the spatial warping path is likely to generate similar segmentation results ($O_c$) as the segmentation path ($S_c$).  The higher the \textit{expected confidence score} is, the more likely the spatial warping path is able to achieve it.  We explain the training methodology of DN in Section~\ref{DN and Its Training Methodology}. Based on the decisions of DN, in step 3 frame regions are forwarded to different paths to generate their regional semantic segmentations.  For the spatial warping path, a special warping function $W(*)$~\cite{DFF} is employed to process the the output of the flow network $F$ with the segmentation $S_k$ from the same region of the key frame to generate a new segmentation $O_c$ for that region.  Please note that the flow network can not generate a regional semantic segmentation by itself.  It simply predicts the displacement of objects by optical flow, and needs to rely on the warping function $W(*)$ and the information contained in the key frames.  We recommend interested readers to refer to~\cite{DFF} for more details of $W(*)$.

\begin{figure}[t]
\centering
\captionsetup{font=small}
\includegraphics[width=0.4\textwidth,height=0.12\textheight]{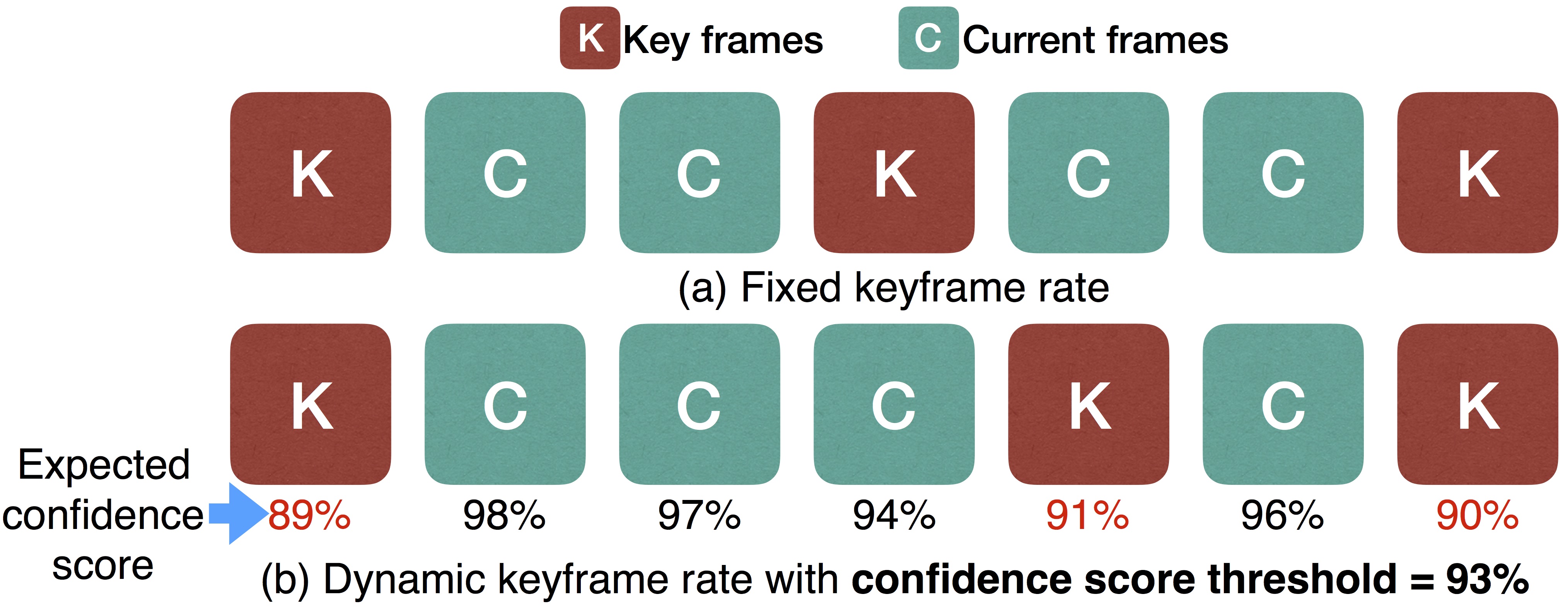}
\vspace{-7pt}
\caption{Different key frame scheduling policies} 
\label{schedule}
\vspace{-18pt}
\end{figure}

\subsection{Adaptive Key Frame Scheduling}
\label{Adaptive Key Frame Scheduling}

Fig.~\ref{schedule} illustrates the key frame scheduling policies used by DFF~\cite{DFF} and DVSNet.  We assume that the sequences of the frame regions in Fig.~\ref{schedule} correspond to the same frame region $r$.  Similar to DFF, DVSNet updates the key frames after a certain period of time.  DFF adopts a fixed update period, as shown in Fig.~\ref{schedule}~(a), which is predetermined and does not take quality and efficiency into consideration.  For example, it is more efficient to process a frame sequence of similar contents with a longer update period, as the spatial warping path itself is sufficient to produce satisfactory outcomes (i.e., regional segmentations).  On the other hand, when the scene changes dramatically, using the segmentation network is more reasonable.  This is because the flow network is unable to predict the displacement of unseen objects not existing in the corresponding key frame region.

\begin{figure}[t]
\centering
\captionsetup{font=small}
\includegraphics[width=0.45\textwidth,height=0.15\textheight]{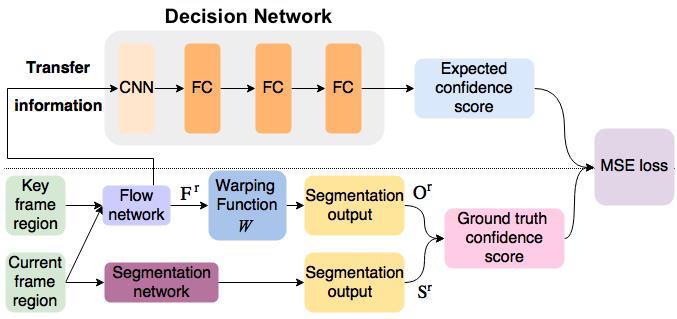}
\vspace{-7pt}
\caption{The network model of DN and its training methodology} 
\label{agent}
\vspace{-15pt}
\end{figure}

We propose an adaptive key frame scheduling policy by using DN and \textit{expected confidence score}.  The adaptive key frame scheduling policy is illustrated in Fig.~\ref{schedule}~(b), in which the update period is not fixed and is determined according to the \textit{expected confidence score} of that region.  DN determines when to update the key frame region $r$ by evaluating if the output of the flow network $F^{r}$ is able to generate a satisfactory regional segmentation $O^{r}$.  If $O^{r}$ is expected to be close to that of the segmentation network $S^{r}$, $F^{r}$ is forwarded to the spatial warping function $W(*)$ to generate $O^{r}$.  Otherwise, the current frame region $I^{r}_{c}$ is sent to the longer segmentation network, and the key frame is updated ($I^{r}_{k}<=I^{r}_{c}$).  Note that DN neither compares $O^{r}$ and $S^{r}$, nor requires them in its evaluation.  It is a regression model trained to "predict" the outcome of $O^{r}$ based on $F^{r}$, as illustrated in Fig~\ref{agent}, and is explained in Section~\ref{DN and Its Training Methodology}. We define a metric, called \textit{confidence score}, to represent the ground truth difference between $ O^{r}$ and $S^{r}$.  Please note that \textit{expected confidence score} and \textit{confidence score} are different.  The former is a value evaluated by DN, while the latter is the ground truth difference in pixels between $ O^{r}$ and $S^{r}$.  The latter is only used in the training phase for training DN, and is not accessible by DN during the execution phase.  The mathematical form of \textit{confidence score} is defined as:
\begin{equation}
\small\small
\vspace{-1.5pt}
confidence\ score = \frac{\sum_{p \in {P}}C({O^{r}}(p),{S^{r}}(p))}{{P}}
\label{equation1}
\vspace{-1.5pt}
\end{equation}
where $P$ is the total number of pixels in $r$, $p$ the index of a pixel, ${O^{r}}(p)$ the class label of pixel $p$ predicted by the spatial warping path, ${S^{r}}(p)$ the class label of pixel $p$ predicted by the segmentation path, $C(u, v)$ a function which outputs 1 only when $u$ equals $v$, otherwise 0.

\begin{figure}[t]
\centering
\captionsetup{font=small}
\includegraphics[width=0.5\textwidth,height=0.2\textheight]{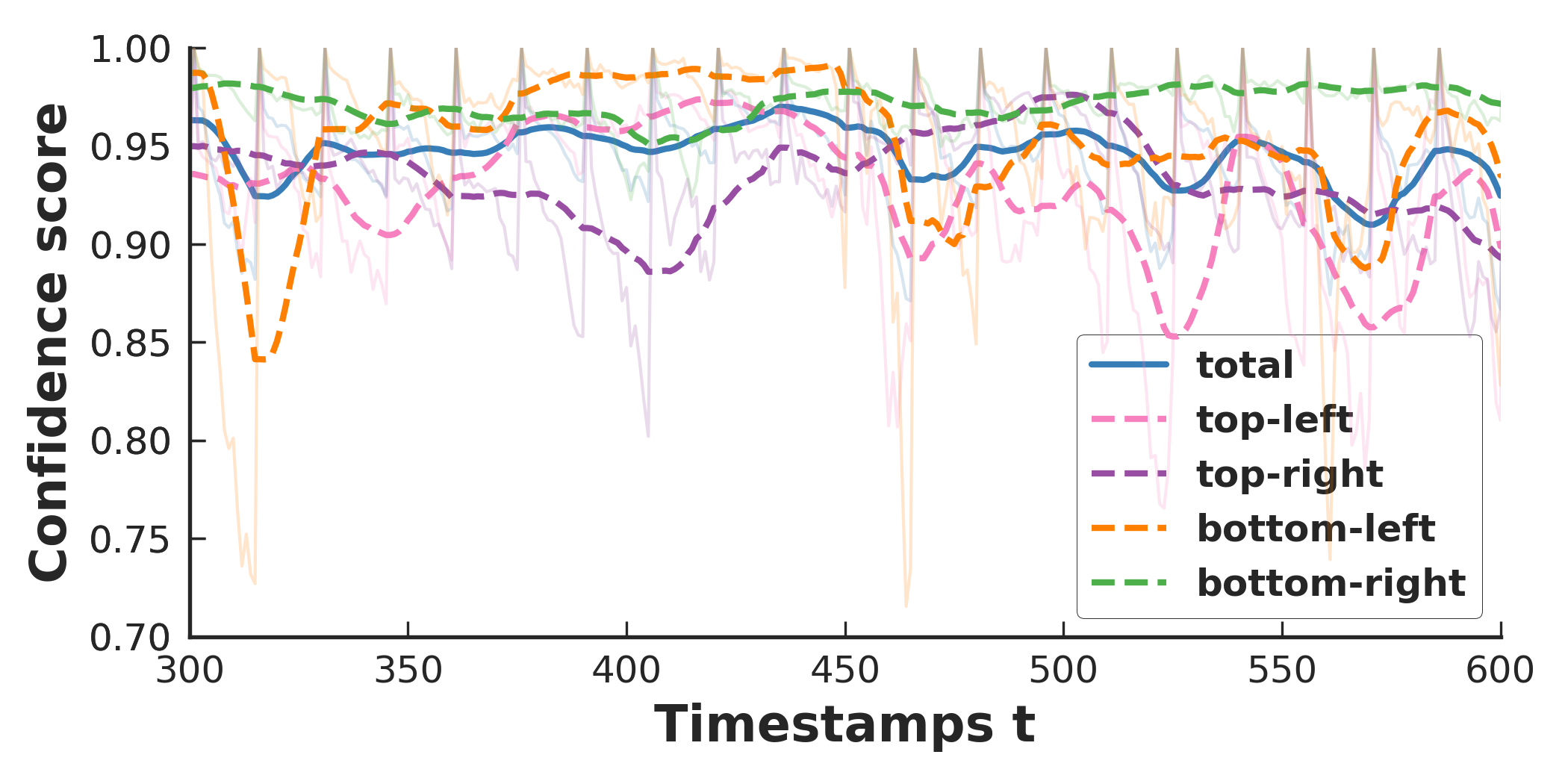}
\vspace{-20pt}
\caption{Confidence score vs. $t$ for the frame regions and the entire frame} 
\label{confidence}
\vspace{-15pt}
\end{figure}

Given a target threshold $t$, DN compares its \textit{expected confidence score} against $t$.  If it is higher than $t$, $F^{r}$ is considered satisfactory.  Otherwise, $I^{r}$ is forwarded to the segmentation path in Fig~\ref{architecture}.  An advantage of the proposed adaptive policy is that the target threshold $t$ is adjustable.  A lower $t$ leads to lower accuracy and higher fps, as more input frame regions traverse the shorter spatial warping path.  On the other hand, a higher $t$ results in higher accuracy, trading off speed for quality.  According to different requirements in different scenes and scenarios, DVSNet can be customized to determine the best $t$ value.

\subsection{Frame Region Based Execution}
\label{Region-Based Execution}
We provide an analytical example to justify the proposed frame region based execution scheme.  Fig.~\ref{confidence} plots curves representing the values of \textit{confidence score} versus time for different frame regions as well as the entire frame for a video sequence extracted from the Cityscape dataset~\cite{Cityscapes}. We again assume that each frame is divided into four frame regions.  We plot the curves from timestamp $300$ to $600$ and use fixed a key frame scheduling policy which updates the key frame every 15 frames.  The curves are smoothed by averaging the data points over 15 timestamps.  The smoothed curves are highlighted in solid colors, while the raw data points are plotted in light colors. It can be seen that the \textit{confidence score} of the entire frame does not fluctuate obviously over time.  However, for most of the time, the \textit{confidence scores} of different frame regions show significant variations.  Some frame regions exhibit high \textit{confidence scores} for a long period of time, indicating that some portions of the frame change slowly during the period. For those scenarios, it is not necessary to feed the entire frame to the segmentation network.  This example validates our claim of using frame region based execution scheme.  In our experimental results, we present a comprehensive analysis for the number of frame regions versus performance.  We also inspect the impact of overlapped pixels between frame regions on DVSNet's accuracy.

\subsection{DN and Its Training Methodology}
\label{DN and Its Training Methodology}
\begin{figure}[t]
\centering
\captionsetup{font=small}
\includegraphics[width=0.4\textwidth,height=0.12\textheight]{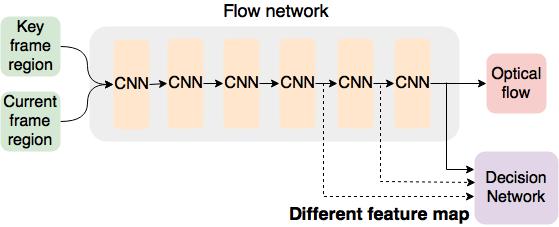}
\vspace{-7pt}
\caption{Different feature maps for training DN} 
\label{feature}
\vspace{-15pt}
\end{figure}

	Fig.~\ref{agent} illustrates the network model of DN as well as its training methodology.  DN is a lightweight CNN consists of only a single convolutional layer and three fully-connected layers.  DN takes as input the feature maps from one of the intermediate layers of the flow network, as illustrated in Fig.~\ref{feature}.  DN is trained to perform regression.  In the training phase, the goal of DN is to learn to predict an \textit{expected confidence score} for a frame region as close to the ground truth \textit{confidence score} (derived from Eq.~(\ref{equation1})) of that region as possible.  The predicted \textit{expected confidence score} is compared with the ground truth \textit{confidence score} to calculate a mean squared error (MSE) loss.  The MSE loss is then used to update the parameters in the model of DN by Adam optimizer~\cite{ADAM}.  In the execution (testing) phase, the ground truth \textit{confidence score} is not accessible to both DN and the flow network.  The feature maps fed into DN is allowed to come from any of the layers of the flow networks, as plotted in Fig.~\ref{feature}.  These feature maps represent the spatial transfer information between a key frame region and its corresponding current frame region~\cite{DFF,Visualize}. We provide an analysis in Section~\ref{Comparison of DN Configurations} for different DN configurations.   

\section{Experiments}
\label{Experiments}
In this section, we present experimental results and discuss their implications.  We start by a brief introduction to our experimental setup in Section~\ref{Experimental Setup}.  Then, we validate our DVSNet for a variety of configurations in Section~\ref{Validation of DVSNet}.
We demonstrate the effectiveness of the proposed adaptive key frame scheduling policy in Section~\ref{Validation of Adaptive Key Frame Scheduling Policy}.  We compare different DN configurations in Section~\ref{Comparison of DN Configurations}.  We analyze different frame division schemes in Section~\ref{Impact of Frame Division Schemes}.  Finally, we evaluate the impact of overlapping frame regions in Section~\ref{Impact of Overlapped Regions on Accuracy}.

\subsection{Experimental Setup}
\label{Experimental Setup}

We perform experiments on the famous Cityscapes dataset~\cite{Cityscapes}. The training, validation, and testing sets contain 2,975, 500, and 1,525 frames, respectively.  The frames of training and validation sets are annotated with pixel-level ground-truth labels for semantic segmentation.  The annotated frame is provided on the $20^{\text{th}}$ frame of a 30-frame video snippet.  We evaluate our approaches on each Cityscapes validation snippet from the $1^{\text{st}}$ frame to the $20^{\text{th}}$ frame.  We set the $1^{\text{st}}$ frame as our initial key frame and measure mIoU on the annotated $20^{\text{th}}$ frames.  

In our experiments, we pre-trained three semantic segmentation models, DeepLab-Fast, PSPNet, and DeepLab-v2, as our baseline models for the segmentation network in DVSNet. DeepLab-Fast is a modified version of DeepLab-v2~\cite{DeepLabv2}, while PSPNet and DeepLab-v2 are reproduced from PSPNet~\cite{PSPNet} and DeepLab-v2~\cite{DeepLabv2}, respectively.  For the reproduced baseline models, we removed several extra features including CRF, multi-scale inferencing, and sliding window segmentation from the original versions to enhance their execution speed.  The values of mIoU and fps of the baseline segmentation models are measured in a per-frame fashion, without any frame division or assistance of the flow network.  The results of the baseline segmentation models on the Cityscape dataset are summarized in Table~\ref{baseline}.  Among the three baseline segmentation models, DeepLab-Fast is three times faster than the other two models, while PSPNet has the highest 77.0\% mIoU accuracy.  We further pre-trained FlowNet2-S and FlowNet2-s to serve as our baseline models for the flow network in DVSNet.  These two models are reproduced from~\cite{FlowNet2}.  In this paper, we represent the DVSNet configuration by a tuple: (\textit{segmentation network, flow network, t}), where $t$ is the target threshold for that DVSNet configuration.   By default, we divide a frame to four regions for the former two models, and two regions for the latter model.  The depth of the overlapped regions between adjacent frame regions is by default set to 64 pixels.  A detailed analysis for justifying this value is provided in Section~\ref{Impact of Overlapped Regions on Accuracy}.  We perform all of our experiments on a server with two Intel Xeon E5-2620 CPUs and an NVIDIA GTX 1080 Ti GPU.

\subsection{Validation of DVSNet}
\label{Validation of DVSNet}

\begin{table}[t]
\centering
\captionsetup{font=small}
\small\small
\begin{tabular}{c|c|c|c}
\hline\hline
\multicolumn{1}{c|}{DVSNet}                      & \multicolumn{1}{c|}{Methods}    & \multicolumn{1}{c|}{mIoU (\%)} & fps \\ \hline\hline
\multicolumn{4}{c}{Baseline segmentation models}                                                                                  \\ \hline
\multicolumn{1}{c|}{\textit{DeepLab-Fast}} & \multicolumn{1}{c|}{per-frame}  & \multicolumn{1}{c|}{73.5}      & 5.6           \\ \hline
\multicolumn{1}{c|}{\textit{PSPNet}}       & \multicolumn{1}{c|}{per-frame}  & \multicolumn{1}{c|}{77.0}      & 1.7           \\ \hline
\multicolumn{1}{c|}{\textit{DeepLab-v2}}    & \multicolumn{1}{c|}{per-frame}  & \multicolumn{1}{c|}{74.8}      & 1.8           \\ \hline\hline
\multicolumn{4}{c}{Balanced mode}                                                                                  \\ \hline
\multicolumn{1}{c|}{\textit{(DeepLab-Fast, FlowNet2-s)}} & \multicolumn{1}{c|}{$t = 92\%$} & \multicolumn{1}{c|}{70.4}      & 19.8           \\ \hline
\multicolumn{1}{c|}{\textit{(PSPNet, FlowNet2-s)}}       & \multicolumn{1}{c|}{$t = 83\%$} & \multicolumn{1}{c|}{70.2}      & 11.5           \\ \hline
\multicolumn{1}{c|}{\textit{(DeepLab-v2, FlowNet2-s)}}    & \multicolumn{1}{c|}{$t = 81\%$} & \multicolumn{1}{c|}{70.3}      & 8.3           \\ \hline
\hline
\multicolumn{4}{c}{High-speed mode}                                                                                                \\ \hline
\multicolumn{1}{c|}{\textit{(DeepLab-Fast, FlowNet2-s)}} & \multicolumn{1}{c|}{$t = 86\%$} & \multicolumn{1}{c|}{63.2}      & 30.4          \\ \hline
\multicolumn{1}{c|}{\textit{(PSPNet, FlowNet2-s)}}       & \multicolumn{1}{c|}{$t = 75\%$} & \multicolumn{1}{c|}{62.6}      & 30.3          \\ \hline
\end{tabular}
\vspace{-8pt}
\caption{Comparison of mIoU and fps for various models, where $t$ represents the target threshold of DN.}
\label{baseline}
\vspace{-15pt}
\end{table}

Table~\ref{baseline} compares the speed (fps) and  accuracy (mIoU) of \textit{(DeepLab-Fast, FlowNet2-s)}, \textit{(PSPNet, FlowNet2-s)}, and \textit{(DeepLab-v, FlowNet2-s)} for two different modes: a balanced mode and a high-speed mode.  The balanced mode requires that the accuracy of a network has to be above 70\% mIoU, while the high-speed mode requires that the frame rate has to be higher than 30 fps. We show the corresponding values of the target threshold $t$ in the $2^{nd}$ column, and the values of mIoU and fps in the $3^{rd}$ and $4^{th}$ columns, respectively.  It is observed that the DVSNet framework is able to significantly improve the performance of the three baseline models.  For the balanced mode, \textit{(DeepLab-Fast, FlowNet2-s, 92)}, \textit{(PSPNet, FlowNet2-s, 83)}, and \textit{(DeepLab-v2, FlowNet2-s, 81)} are 3.5$\times$, 6.8$\times$, and 4.6$\times$ faster than their baseline counterparts, with a slight drop of 3\%, 6\%, and 4\% in mIoU, respectively.  The high-speed mode enables the two models to achieve real-time speed. The frame rates of \textit{(DeepLab-Fast, FlowNet2-s, 86)} and \textit{(PSPNet, FlowNet2-s, 75)} are 5.4$\times$ and 17.8$\times$ faster than their baseline counterparts, respectively.  The mIoU of them declines to 63.2\% and 62.6\%, respectively.  From Table~\ref{baseline}, we conclude that decreasing $t$ leads to a drop in mIoU of the models, but increases fps significantly.  

\begin{figure}[t]
\centering
\captionsetup{font=small}
\includegraphics[width=0.45\textwidth,height=0.2\textheight]{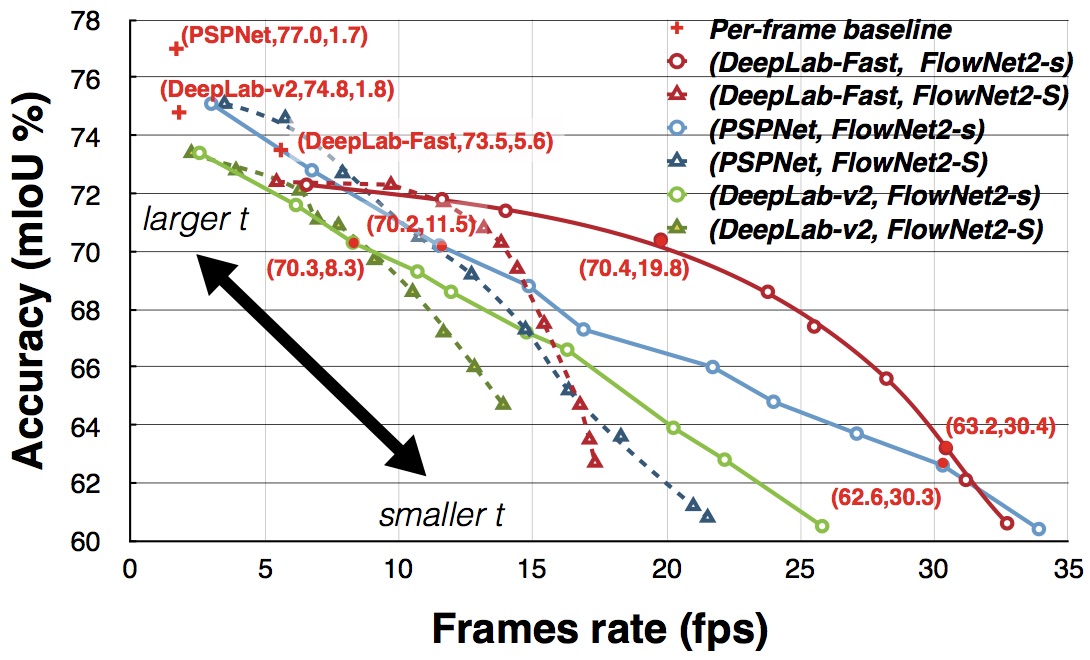}
\vspace{-10pt}
\caption{Accuracy (mIoU) and frame rate (fps) of various DVSNet configurations under different threshold $t$.} 
\label{fps}
\vspace{-17pt}
\end{figure}

Fig.~\ref{fps} shows accuracy (mIoU) versus frame rate (fps) for various DVSNet configurations.  We plot six curves on Fig.~\ref{fps}, corresponding to six possible combinations of the three baseline segmentation network models and the two baseline flow network models.  The three solid curves represent DVSNet configurations built on top of FlowNet2-s, while the remaining three dashed curves stand for DVSNet configurations using FlowNet2-S.  The data points on each curve correspond to different values of target threshold $t$ under the same DVSNet configuration.  It can be observed that as $t$ increases, the data points of all curves move toward the upper-left corner, leading to increased mIoU accuracies but decreased fps for all DVSNet configurations.  On the contrary, when $t$ decreases, the data points of all curves move toward the opposite bottom-right corner, indicating that more frame regions pass through the shorter spatial warping path.  It can be seen that the dashed lines drop faster than the solid lines, because FlowNet2-S is deeper and thus slower than FlowNet2-s.  By adjusting the value of $t$ and selecting the baseline models, DVSNet can be configured and customized to meet a wide range of accuracy and frame rate requirements.

\subsection{Validation of DVSNet's Adaptive Key Frame Scheduling Policy}
\label{Validation of Adaptive Key Frame Scheduling Policy}

\begin{figure}[t]
\centering
\captionsetup{font=small}
\includegraphics[width=0.45\textwidth,height=0.2\textheight]{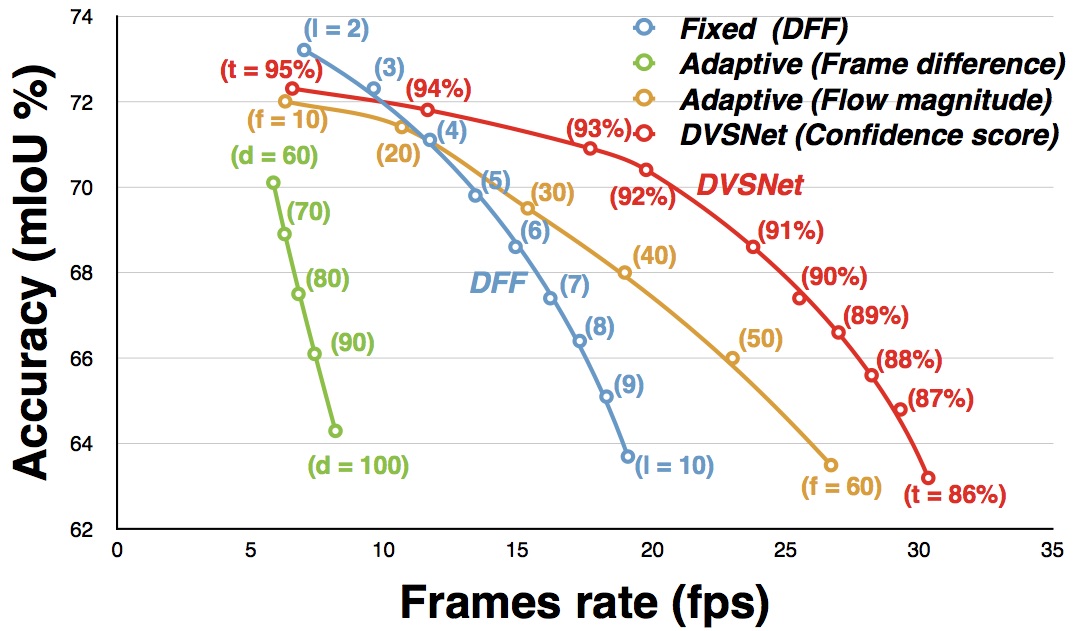}
\vspace{-11pt}
\caption{Accuracy (mIoU) versus frame rate (fps) under different key frame scheduling policies. $t$ is the target confidence score threshold. $l$ is the key frame update period in DFF~\cite{DFF}. $d$ is the frame difference threshold.  $f$ is the flow magnitude threshold.} 
\label{fix2adapt}
\vspace{-16pt}
\end{figure}

In this section, we validate the effectiveness of DVSNet's adaptive key frame scheduling policy.  Fig.~\ref{fix2adapt} plots a comparison of performance between the fixed key frame scheduling policy and the adaptive key frame scheduling policy.  DVSNet which adopts adaptive scheduling with \textit{expected confidence score} and DFF~\cite{DFF} which adopts fixed scheduling correspond to the red and blue curves, respectively.  The evaluation method of the fixed key frame scheduling policy employed by DFF tends to overestimate mIoU. For each frame $i$ with ground truth annotation, DFF averages mIoU from $m$ image pairs $(k, i)$, where key frames $k = i-(m+1), ..., i$. The method seems counter-intuitive and fails to reflect the impact of scheduling policies on mIoU.  In our fixed key frame scheduling policy experiment, we start from the key frames $k = i-(l+1)$, where $l$ is key frame update period, and measure mIoU at each frame $i$. 

We include additional curves in Fig.~\ref{fix2adapt} to compare the impact of another two different decision metrics for the adaptive key frame scheduling policy: frame difference (green curve) and flow magnitude (orange curve).  The frame difference $d$ and the flow magnitude $f$ are given by:
\begin{equation}
\small\small
d =\frac{\sum_{p \in {P}}(G(I_{k})_{p}-G(I_{i})_{p})}{P}
\label{equation2}
\end{equation}
\begin{equation}
\small\small
f =\frac{\sum_{p \in {P}}(\sqrt{u_{p}^{2}+v_{p}^{2}})}{P}
\label{equation3}
\end{equation}
where $P$ is the total number of pixels in a frame or frame region, $p$ represents the index of a pixel, $G(*)$ is a grayscale operator which converts an RGB image to a grayscale one, and $u$ and $v$ represent the horizontal and vertical movements, respectively.  For a fair comparison, the networks are all configured to~\textit{(DeepLab-Fast, FlowNet2-s)}.  

Fig.~\ref{fix2adapt} reveals that using frame difference as the decision metric for the adaptive key frame scheduling policy is inadequate. For all values of $d$ considered in our experiments (ranging from 60 to 100), the mIoU accuracies and the frame rates of the datapoints on the green curve are much lower than those of the other three curves.  On the other hand, it is observed that using the adaptive key scheduling policy with either \textit{expected confidence score} or flow magnitude as the decision metrics deliver higher mIoU accuracies than the fixed key frame scheduling policy employed by DFF, even at high fps.  The curves indicate that DVSNet employing the adaptive scheduling policy with \textit{expected confidence score} as the decision metric results in the best performance.  This experiment validates the effectiveness of DVSNet's adaptive key frame scheduling policy.

\subsection{Comparison of DN Configurations}
\label{Comparison of DN Configurations}

We perform experiments for a number of DN configurations illustrate in Fig.~\ref{feature}, and summarize the results in Table~\ref{featuret}.  We feed DN with feature maps from different layers of the flow network, including the feature maps after the $4^{th}$, $5^{th}$, and $6^{th}$ convolutional layers.  We additional include an entry in Table~\ref{featuret} to represent the case where DN is fed with the key frame and the output of the flow network.  We measure the performance of these variants by comparing the error rate of the evaluated \textit{expected confidence score} against the ground truth \textit{confidence score}.  It is observed that the feature maps after the $6^{th}$ convolutional layers lead to the lowest error rate.  

\begin{table}[t]
\centering
\captionsetup{font=small}
\small\small
\begin{tabular}{c|c}
\hline\hline
Inputs fed to DN           & Error rate(\%) \\ \hline\hline
key-frame + flow & 2.20       \\ \hline
Feature maps after conv4            & 1.94       \\ \hline
Feature maps after conv5            & 1.85       \\ \hline
Feature maps after conv6            & \textbf{1.74}       \\ \hline
\end{tabular}
\vspace{-8pt}
\caption{Impact of input feature maps on DN's performance}
\label{featuret}
\vspace{-10pt}
\end{table}

\begin{table}[t]
\centering
\captionsetup{font=small}
\small\small
\begin{tabular}{c|c}
\hline\hline
Dividing Methods & \multicolumn{1}{c}{Spatial Warping / Segmenation} \\ \hline\hline
Original     & $4,535 / 5,465 = 0.83$                                \\ \hline
Half         & $9,798 / 10,202 = 1.04$                               \\ \hline
$2 \times 2$ & $34,013 / 5,987 = \textbf{5.68}$                      \\ \hline
$3 \times 3$ & $69,157 / 20,843 = 3.32$                              \\ \hline
$4 \times 4$ & $122,958 / 37,042 = 3.32$                             \\ \hline
\end{tabular}
\vspace{-8pt}
\caption{Impact of frame division schemes on performance}
\label{divide}
\vspace{-17pt}
\end{table}

\subsection{Impact of Frame Division Schemes}
\label{Impact of Frame Division Schemes}
Table~\ref{divide} shows the impact of different frame division schemes on performance.  Table~\ref{divide} compares five different schemes corresponding to five different ways to divide a video frame.  The ratio of the  number of frame regions sent to the spatial warping path to the number of those sent to the segmentation path is represented as $Spatial\ Warping/Segmentation$.  A higher ratio indicates that more frame regions are forwarded to the spatial warping path.  We observe that the $2\times 2$ division scheme offers the best utilization of the spatial warping path.  The results also show that too few or too many frame regions do not exploit the spatial warping path well.  A frame without any frame regions prohibits DVSNet from exploiting non-homogeneity in frame differences between consecutive frames.  Too many frame regions lead to large variations in \textit{expected confidence score}, causing many of them to be forwarded to the segmentation network frequently.

\begin{figure}[t]
\centering
\captionsetup{font=small}
\includegraphics[width=0.45\textwidth,height=0.19\textheight]{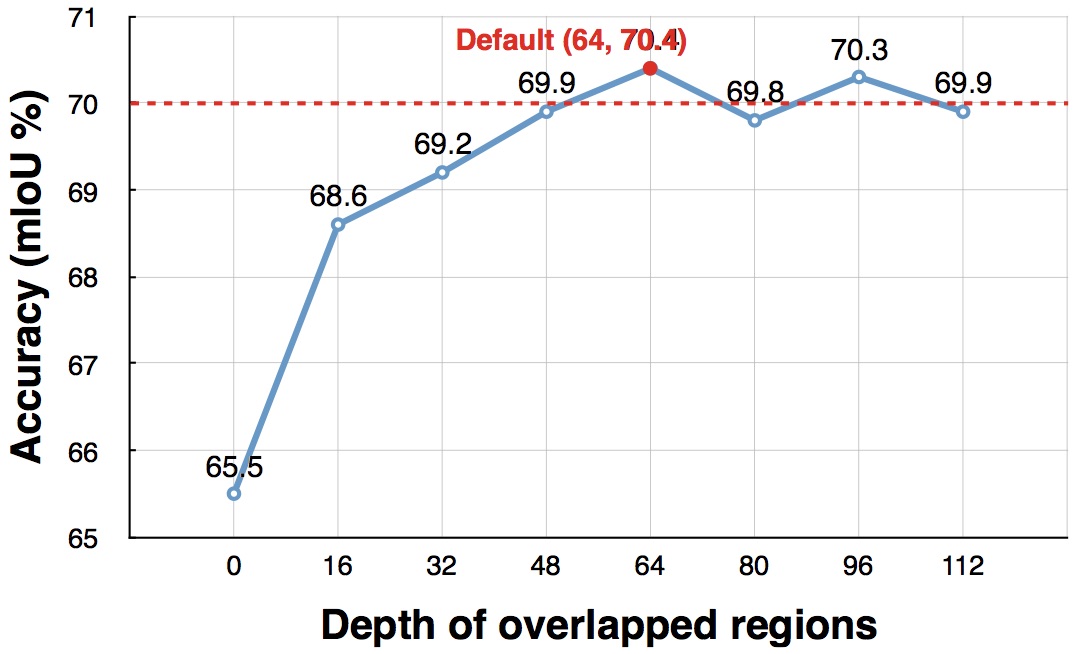}
\vspace{-8pt}
\caption{Impact of overlapping frame regions on accuracy} 
\label{overlap}
\vspace{-15pt}
\end{figure}

\subsection{Impact of Overlapped Regions on Accuracy}
\label{Impact of Overlapped Regions on Accuracy}
Fig~\ref{overlap} shows the accuracy of DVSNet as a function of the depth of overlapped frame regions. We assume that each video frame is divided to four frame regions. The configuration of DVSNet is set to be \textit{(DeepLab-Fast, FlowNet2-s, 92)}. The x-axis represents the overlapped depth in pixels between adjacent frame regions.  The y-axis represents the mIoU accuracy.  Without overlapped regions, DVSNet is only able to achieve an accuracy up to 65.5\% mIoU.  However, when the depth of the overlapped region increases, the mIoU accuracy also increases.  The curve saturates at 70\% mIoU, corresponding to an overlapped depth of 64 pixels.  However, more overlap pixels lead to larger frame regions, resulting in increased computation time.  According to our experiments, an empirical value of 64 pixels seem to be optimal for DVSNet.  We plot a red dotted line corresponding to the mIoU accuracy of the entire frame without any frame division in Fig~\ref{overlap} for the purpose of comparison.

\section{Conclusion}
\label{Conclusion}
We presented a DVSNet framework to strike a balance between quality and efficiency for video semantic segmentation.  The DVSNet framework consists of two major parts: a segmentation path and a spatial warping path.  The former is deeper and slower but highly accurate, while the latter is faster but less accurate. We proposed to divide video frames into frame regions, and perform semantic segmentation for different frame regions by different DVSNet paths.  We explored the use of DN to determine which frame regions should be forwarded to which DVSNet paths based on a metric called \textit{expected confidence score}.  We further proposed an adaptive key frame scheduling policy to adaptively adjust the update period of key frames at runtime.  Experimental results show that DVSNet is able to achieve up to 70.4\% mIoU at 19.8 fps on the Cityscape dataset.  We have performed extensive experiments for various configurations of DVSNet, and showed that DVSNet outperforms contemporary state-of-the-art semantic segmentation models in terms of efficiency and flexibility.  

\section{Acknowledgments}
\label{Acknowledgments}
The authors thank MediaTek Inc. for their support in researching funding, and NVIDIA Corporation for the donation of the Titan X Pascal GPU used for this research.


{
\small\small
\bibliographystyle{ieee}
\let\oldbibliography\thebibliography
\renewcommand{\thebibliography}[1]{%
  \oldbibliography{#1}%
  \setlength{\itemsep}{0pt}%
}

\end{document}